\documentclass[a4paper,twoside]{article}

\usepackage{epsfig}
\usepackage{subcaption}
\usepackage{calc}
\usepackage{amssymb}
\usepackage{amstext}
\usepackage{amsmath}
\usepackage{amsthm}
\usepackage{multicol}
\usepackage{pslatex}
\usepackage{apalike}
\usepackage{algorithm2e}
\usepackage[bottom]{footmisc}

\usepackage[inline]{enumitem}
\usepackage{hyperref}
\usepackage[nohyperlinks, nolist]{acronym}

\usepackage{listings}
\usepackage{xcolor}

\usepackage[]{todonotes}

\usepackage{SCITEPRESS}     % Please add other packages that you may need BEFORE the SCITEPRESS.sty package.
\newif\ifblinded
\newif\ifnopackagenames
\blindedfalse   % camera-ready
\nopackagenamesfalse % package names are okay
\newcommand{\reffig}[1]{Figure ~\ref{#1}}
\newcommand{\refsec}[1]{Section ~\ref{#1}}

\begin{document}
\title{\LARGE \bf Making Robots Play by the Rules: The ROS 2 CLIPS-Executive}

\ifblinded
\author{
  % keep space but hide content
  Anonymous Author(s)
  \phantom{Tarik Viehmann, Daniel Swoboda, Samridhi Kalra, Himanshu Grover and Gerhard Lakemeyer}
  \vspace{1em}
}
\else
\author{
	\authorname{Tarik Viehmann\sup{1}\orcidAuthor{0000-0003-0264-0055}, Daniel Swoboda\sup{2}\orcidAuthor{0000-0002-3189-3089}, Samridhi Kalra\sup{2}\orcidAuthor{0009-0002-3044-959X}, Himanshu Grover\sup{3}\orcidAuthor{0009-0009-7109-6218} and Gerhard Lakemeyer\sup{1}\orcidAuthor{0000-0002-7363-7593}}
\affiliation{\sup{1}Knowledge-Based Systems Group, RWTH Aachen University, Aachen, Germany}
\affiliation{\sup{2}Chair of Machine Learning and Reasoning (i6), RWTH Aachen University, Aachen, Germany}
\affiliation{\sup{3}MASCOR Institute, FH Aachen University of Applied Sciences, Aachen, Germany}
\email{viehmann@kbsg.rwth-aachen.de, daniel.swoboda@ml.rwth-aachen.de}
}
\fi
\keywords{ROS, CLIPS, Reasoning, Cognitive Robotics, PDDL, Automated Planning}

\abstract{%The abstract should summarize the contents of the paper and should contain at least 70 and at most 200 words. The text must be set to 9-point font size.
CLIPS is a rule-based programming language for building knowledge-driven applications, well suited for the complex task of coordinating autonomous robots.
Inspired by the CLIPS-Executive originally developed for the lesser known Fawkes robotics framework, we present an Integration of CLIPS into the ROS ecosystem.
Additionally, we show the flexibility of CLIPS by describing a PDDL-based planning framework integration.
}

\onecolumn \maketitle \normalsize \setcounter{footnote}{0} \vfill

\begin{acronym}
\acro{CLIPS}{C-Language Integrated Production System}
\acro{LHS}{left-hand side}
\acro{RHS}{right-hand side}
\acro{ROS}{Robot Operating System}
\acro{PDDL}{Planning Domain Definition Language}
\acro{RCLL}{RoboCup Logistics League}
\acro{STN}{Simple Temporal Network}
\acro{CX}{CLIPS-Executive}
\acro{UPF}{Unified Planning Framework}
\end{acronym}
\newcommand\anonlink{\footnote{anonymized for review} }

%%%%%%%%%%%%%%%%%%%%%%%%%%%%%%%%%%%%%%%%%%%%%%%%%%%%%%%%%%%%%%%%%%%%%%%%%%%%%%%%

%%%%%%%%%%%%%%%%%%%%%%%%%%%%%%%%%%%%%%%%%%%%%%%%%%%%%%%%%%%%%%%%%%%%%%%%%%%%%%%%
\section{\uppercase{Introduction}}
\label{sec:introduction}
In recent years, the \ac{ROS} 2~\cite{macenski_robot_2022} developed into the de-facto standard robotics framework, providing a large selection of standard tools and a sophisticated feature set for core robotics tasks, such as communication, hardware abstraction, visualization and simulation tools and navigation.
However, while the ROS 2 ecosystem offers many solutions to develop core robotic skills, additional challenges arise, when these skills have to be used in the context of robotic applications.
Autonomous robots are expected to choose their actions intelligently, react to dynamic changes in the environment flexibly and also coordinate with other robots or humans.
A promising approach to address these requirements is to utilize knowledge-based reasoning, which allows to operate on symbolic knowledge, model domain-specific constraints, and support decision-making at multiple levels of abstraction.
\ac{CLIPS}, originally developed at NASA’s Lyndon B. Johnson Space Center, provides a rule-based inference engine well-suited for building knowledge-driven applications. While it has been widely used in query-based recommendation systems across diverse domains~\cite{miron_flickers_2019,matelli_development_2011}, it is also a suitable framework for reasoning and coordination in robotics applications~\cite{savage_semantic_2019,cengeloglu_dynaclips_1994}.
One such example is the \emph{Fawkes} robotics framework~\cite{niemueller_design_2010}, which deeply integrates \ac{CLIPS} with other robotics components~\cite{niemueller_goal_2019}, enabling users to configure and use different \ac{CLIPS} environments as well as providing the ability to inject external feedback and knowledge through custom plugins.

Following this line of work, we present the ROS 2 \ac{CX}, an integration of CLIPS to build knowledge-driven applications for ROS.
%It features a \ac{ROS} node to manage \ac{CLIPS} environments, a plugin system for extending \ac{CLIPS} with custom features, as well as a collection of plugins for loading of \ac{CLIPS} code, continuous execution and interoperability with \ac{ROS} (e.g., communication via topics, services and actions).
%This allows users to leverage CLIPS-based reasoning in a broad range of applications that benefit from reactive control and symbolic knowledge representation in a novel, ground-up implementation designed specifically for the ROS~2 ecosystem.
The specific contributions of this paper are:

% \begin{itemize}
%     \item \textbf{A complete redesign of the \ac{CX}} as a modern ROS~2 Lifecycle Node. Our framework synthesizes reactive CLIPS-based reasoning with deliberative PDDL planning and standard ROS~2 subsystems, establishing a robust architecture for high-level robot control and agent development.
%     \item  \textbf{A novel plugin system} that decouples the reasoning engine from system and robot specific capabilities.
%     \item \textbf{A novel bridge between the rule-based execution of CLIPS and the Unified Planning Framework (UPF)}. This integration allows the executive to dynamically manage PDDL domains and dispatch temporal actions while retaining the reactivity of a rule-based system.
%     \item \textbf{Our framework as open-source software}, including a verified vendor package for CLIPS, a suite of standard interaction plugins (ROS topics, services, actions), and the PDDL management tools, providing the community with a ready-to-use reasoning backend.
% \end{itemize}
% Tarik: As it's core, the ROS CX is not about PDDL and I would also rather say it is a new framework for ROS that is inspired by an implementation
\begin{itemize}
    \item \textbf{A novel CLIPS reasoning system for ROS.} Designed as the spiritual successor to the Fawkes-based CLIPS integration, this framework enables reactive reasoning in ROS while emphasizing general-purpose functionality and minimal configuration effort.
    \item \textbf{A dynamic plugin system.} The reasoning engine is decoupled from system- and robot-specific capabilities via a flexible plugin architecture. Several standard plugins enhance classical CLIPS reasoning with features for interacting with the ROS ecosystem (e.g., interfacing via topics, services and actions).
    \item \textbf{A novel approach to combining CLIPS and PDDL planning.} As an extension to the core \ac{CX} feature set, we provide an external component for PDDL-based reasoning leveraging the \ac{UPF}. This integration allows the executive to manage PDDL domains and dispatch temporal actions, preserving the reactivity of a rule-based system alongside the capabilities of dedicated planning frameworks.
    \item \textbf{Open-source availability.} Our framework is released as open-source software
\ifblinded
under the Apache 2.0 license\anonlink and binaries are also available\anonlink.
\else
under the Apache 2.0 license\footnote{\url{https://github.com/carologistics/clips_executive}}, and binaries are also available\footnote{\url{https://index.ros.org/p/clips_executive/}}.
\fi
, providing the community with a ready-to-use and well documented reasoning backend.
\end{itemize}

%Utilizing the ROS interoperability features of the \ac{CX}, we demonstrate the benefits of combining CLIPS-based reasoning with specialized decision-making frameworks, by presenting a \ac{PDDL} integration utilizing the \ac{UPF}~\cite{micheli_unified_2025}.
%As a real-world application example, we consider the \ac{RCLL}, where we describe an agent system centered around the \ac{CX} and \ac{PDDL} planning.
Additionally, we report on our efforts to deploy the presented framework in a real-world scenario, the \ac{RCLL}, where we constructed an agent system built around the \ac{CX} and PDDL-based planning.

\section{\uppercase{Related Work}}
% Closest to our work is the Fawkes \ac{CLIPS} integration, which fostered the development of various autonomous robotic systems over the last decade~\cite{niemueller_incremental_2013,niemueller_goal_2019,hofmann_multi-agent_2021,swoboda_towards_2022}.
% Its functionality is similar to the presented implementation in that \ac{CLIPS} serves as an executive component, enabling continuous reasoning as well as enabling interaction with other system components.
% However, there are significant differences: Whereas Fawkes provides a monolithic robotics framework with its own middleware, our approach integrates \ac{CLIPS} directly into the \ac{ROS}~2 ecosystem, thereby leveraging its widespread adoption and interoperability.
% In contrast to the tightly coupled Fawkes integration, the \ac{CX} is realized as a modular \ac{ROS} node with a plugin system, making it reusable across diverse applications.
While conceptually inspired by the Fawkes CLIPS-Executive~\cite{niemueller_incremental_2013,niemueller_goal_2019,hofmann_multi-agent_2021,swoboda_towards_2022}, the \ac{CX} is not a port of the Fawkes codebase. The legacy Fawkes implementation relied on the clipsmm library and a custom middleware layer. In contrast, the CX is a standalone implementation that directly utilizes the new C++ API introduced in CLIPS 6.4 and adopts modern ROS 2 design patterns, such as Lifecycle management and composable nodes. This modernizes the executive from a specialized component within a legacy framework to a general-purpose, native ROS 2 application. Some of the Fawkes implementations' key additions (like PDDL integration) are also part of our framework, but now leveraging the modularity of ROS to rely on dedicated frameworks instead of integrating functionalities directly in the executive in a monolithic way.
%with modern interfacing and in a ROS2 native way.

% Moreover, our work goes beyond the capabilities of Fawkes by enabling seamless interaction with \ac{ROS} communication primitives.
% An initial attempt to migrate the Fawkes \ac{CLIPS}-based Goal Reasoning framework~\cite{niemueller_goal_2019} to \ac{ROS} was presented in earlier
% \ifblinded unpublished work~\cite{anonymousbc},
% \else unpublished work~\cite{doychev21},
% \fi
% which provided a proof of concept but lacked direct integration with \ac{ROS}.
% The current system represents a complete redesign that generalizes and modernizes the original approach.
% Rather than being a direct port of the Fawkes components, the \ac{CX} constitutes a standalone and contemporary successor that lifts the CLIPS-Executive from a specialized component in a legacy framework to a fully integrated \ac{ROS} application.
%It directly vendors the latest \ac{CLIPS} version, employs the revised \ac{CLIPS} API introduced in version~6.4.x instead of relying on \texttt{clipsmm} as an external bridge, and adopts modern~C++ practices such as smart pointers for robust memory management.

Moreover, our work extends beyond Fawkes by enabling seamless interaction with \ac{ROS} communication primitives.
An initial migration of the Fawkes \ac{CLIPS}-based Goal Reasoning framework~\cite{niemueller_goal_2019} to \ac{ROS} was explored in earlier
\ifblinded
unpublished work~\cite{anonymousbc},
\else
unpublished work~\cite{doychev21},
\fi
which was later rewritten to achieve a more general and maintainable design and without a fixation on goal reasoning as a mechanism to enable all kinds of CLIPS-based reasoning without restriction.
We consider the presented system architecturally derived from the CLIPS integration of Fawkes, lifting the CLIPS-Executive from being a specialized component in a legacy framework to becoming a standalone and more generalized ROS application.
On a technical level, the presented system provides several advantages: It directly vendors the most recent \ac{CLIPS} version, leverages the revised \ac{CLIPS} API introduced in version 6.4  and adopts modern C++ practices such as smart pointers for memory management.

%There are also a few other frameworks available that offer knowledge-based systems for high-level control and reasoning in ROS.
% Existing frameworks for knowledge-based reasoning and high-level control in ROS rely on different paradigms.
% PlanSys2~\cite{martin_plansys2_2021} is centered around PDDL-based planning and plan execution. It transforms obtained plans into behavior trees for execution.
% In contrast, the \ac{CX} is not limited to planning, but provides a more general reasoning framework built around \ac{CLIPS}. Within this framework, PDDL planning can be integrated as one reasoning mechanism among others.
% \todo{maybe we can add that as much or as little reasoning as desired can be done within CLIPS}
% For example, PlanSys2 can serve as a backend planner, while the \ac{CX} leverages its results in combination with rule-based inference and external knowledge sources. This allows developers to benefit from the structured action models and plan generation of PlanSys2, while retaining the flexibility of continuous knowledge-driven reasoning in the \ac{CX}.
% We present a similar approach, but using a different PDDL backend, in \refsec{sec:PDDL-CLIPS}
Existing frameworks for knowledge-based reasoning and high-level control in ROS rely on various different paradigms.
PlanSys2~\cite{martin_plansys2_2021} is centered around PDDL-based planning and plan execution, transforming obtained plans into behavior trees for execution.
In contrast, the \ac{CX} is not limited to planning, but provides a more general reasoning framework built around \ac{CLIPS}. Within this framework, PDDL planning can be integrated as one reasoning mechanism among others. For example, PlanSys2 can serve as a planner, while the \ac{CX} leverages its results in combination with rule-based inference and external knowledge sources.
This enables developers to decide how much reasoning they wish to realize inside \ac{CLIPS} (from minimal monitoring rules to extensive agent-specific policies) while still benefiting from the structured action models and plan generation of PlanSys2.
We demonstrate a similar approach using a different PDDL backend in \refsec{sec:PDDL-CLIPS}.

% SkiROS2~\cite{mayr_skiros2_2023} is a frameworks for building skills using behaviour trees. Skills are annotated with ontology-based descriptions providing conditions and effects. Additionally, a task manager is used to perform PDDL-based planning, by first transforming the skill descriptions and current knowledge to PDDL, then invoking the TFD planner and finally transforming the resulting plan to a behavior tree that can be executed by the skill manager.
% While ontologies provide a structured representation of skill preconditions and effects, they are comparatively rigid and limited in expressiveness. CLIPS rules, on the other hand, allow dynamic inference, procedural knowledge, and reactive adaptation, enabling richer reasoning during execution.
SkiROS2~\cite{mayr_skiros2_2023} is a framework for building skills using behavior trees. Skills are annotated with ontology-based descriptions that define their conditions and effects. A task manager enables PDDL-based planning by transforming the skill descriptions and current knowledge into PDDL, invoking the TFD planner, and converting the resulting plan into a behavior tree executable by the skill manager.
Ontologies offer a well-structured and semantically grounded representation of skill preconditions and effects, which is advantageous for interoperability and knowledge sharing. However, they typically emphasize declarative descriptions and are less suited for encoding procedural or reactive knowledge. In contrast, \ac{CLIPS} rules naturally support dynamic inference, context-dependent reasoning, and reactive adaptation at runtime, enabling richer forms of reasoning during execution.

% ROSA~\cite{silva_rosa_2025} is a Task and Architecture Co-Adaptation (TACA) framework that uses TypeDB to store and update the knowledge base. Heuristic Rules are defined using TypeQL in order to adapt the tasks and architecture to dynamic changes. ROSA also implements executor interfaces to PlanSys2 and to BehaviorTreeCPP\footnote{\url{https://www.behaviortree.dev/}}.
% While TypeQL provides a powerful query language for structured knowledge graphs, its reasoning capabilities are limited to query evaluation and explicit updates.
% Moreover, ROSA relies on a fixed schema and enforces specific entity relationships to ensure consistency. This makes it less flexible for procedural or reactive knowledge representation. In contrast, the \ac{CX} imposes no additional modeling constraints beyond those inherent to \ac{CLIPS}, allowing developers to encode arbitrary rules and facts without schema restrictions, and to combine declarative and procedural reasoning in a single framework.
ROSA~\cite{silva_rosa_2025} is a Task and Architecture Co-Adaptation (TACA) framework that uses TypeDB to maintain and update its knowledge base. Heuristic rules are expressed in TypeQL, which supports adapting both tasks and system architecture to dynamic changes. ROSA also implements executor interfaces to PlanSys2 and to BehaviorTreeCPP\footnote{\url{https://www.behaviortree.dev/}}.
TypeQL offers a powerful query language for structured knowledge graphs, particularly well-suited to managing consistency and explicit relationships between entities. However, its reasoning capabilities are centered on query evaluation and rule-based updates within a fixed schema, making it less flexible for procedural or reactive forms of knowledge. In contrast, the \ac{CX} imposes no modeling constraints beyond those inherent to \ac{CLIPS}, enabling developers to encode arbitrary rules and facts without schema restrictions, and to combine declarative and procedural reasoning in a single framework.

%Recent work in general-purpose service robotics has explored deep learning and LLMs for multi-task skill execution.
%While multi-task pretraining improves skill execution and generalization, the relatively low success rates on complex multi-step tasks indicate that this approach is not yet a fully reliable solution for general-purpose service robotics~\cite{lbmtri2025}.
In another line of work symbolic reasoning provides a structural framework to LLM-based behaviour.
Contreras et al.~\cite{contreras_complex_2025} combine conceptual dependency parsing with a CLIPS-based expert system to generate structured, optimal plans that handle missing information and dynamically execute skills.
However, they tightly integrate CLIPS into their application. A more general framework for CLIPS like the \ac{CX} may foster its usage in more applications.

 In summary, while existing ROS frameworks rely on PDDL, ontologies, or specialized databases to deal with structured knowledge, our framework leverages a rule-based production system (CLIPS) that naturally unifies knowledge representation and reasoning with execution.
 In contrast to systems that implement specific planning or execution paradigms, our approach provides a general-purpose expert system environment that freely interacts with ROS 2 in the reasoning loop, enabling both reactive and deliberative control without translation overhead.

% In summary, existing ROS~2 frameworks each emphasize a particular paradigm:
% PlanSys2 focuses on plan generation and behavior tree execution,
% SkiROS2 couples skills with ontologies and PDDL planning,
% and ROSA relies on schema-based knowledge graphs with heuristic adaptation.
% Recent work, such as that by Contreras et al., demonstrates the value of combining language models and CLIPS, but only in a task-specific way.
% The CX differs by combining many of their strengths within a single, reusable framework: it supports structured plan generation via external planners,
% semantic-style reasoning through CLIPS rules, reactive adaptation to exogenous events,
% and integration with behavior tree or other execution engines.
% Developers can choose how much reasoning to realize inside CLIPS---from lightweight monitoring to complex agent policies---while retaining the benefits of PDDL planning and many other types of reasoning through standardized ROS interfaces.

\section{\uppercase{Background}}
This section provides the necessary background for the proposed \ac{CX} framework.
First, we introduce the \ac{CLIPS} rule-based system, which forms the foundation of \ac{CX}.
Then, we discuss \ac{PDDL}, the standardized representation language for modern planners, which plays a key role in our real-world application described in \refsec{sec:PDDL-CLIPS}.

\subsection{\uppercase{ROS 2}}\label{sec:ROS}
ROS~2~\cite{macenski_robot_2022} has become the de facto standard middleware in robotics, providing a unified communication infrastructure and a rich ecosystem of first and third party libraries for relevant robotics skills like perception, control, navigation, and simulation.
The functionality is wrapped in nodes that use standardized communication interfaces to interact. Its architecture builds on a publisher and subscriber paradigm with additional abstractions for services and actions. ROS~2 additionally introduces the concept of lifecycle management which allows nodes to be configured and monitored. 

\subsection{\uppercase{CLIPS}}\label{sec:CLIPS}
\begin{figure}
\includegraphics{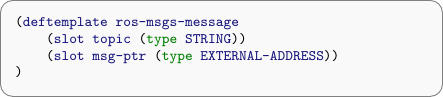}
\caption{Example CLIPS deftemplate representing an external ROS message.}
\label{fig:deftemplate}
\end{figure}

\ac{CLIPS} is a rule-based programming language designed for building expert systems. It was developed in the 1980s by NASA for developing knowledge-based applications.
Its documentation, including syntax, semantics and API, can be found on the official website\footnote{\url{https://clipsrules.net}}.
CLIPS programs center around facts, object-like structured information stored in a knowledge base. Each fact comprises of basic data types such as integers, floats, external memory addresses, or strings). An example of how to define a new fact template is depicted inf \reffig{fig:deftemplate}, where facts of type \texttt{ros-msgs-message} are defined to have two attributes (called \emph{slots}), one containing a string, and one holding external data.
Heuristic knowledge in the form of user-defined rules specify how the fact list can evolve, and thus how a process might be controlled given the current state of the knowledge base.
A rule is comprised of an \emph{antecedent} (or left-hand side, LHS) providing conditions to the fact list under which it can be activated, as well a \emph{consequent} (or right-hand side, RHS) which define the effects when a rule is \emph{fired} (meaning being executed).
Different scopes are always encapsulated by parentheses (e.g., facts, functions, rules, slots). The LHS and RHS are separed by an arrow $=>$ for better readability instead.
Rule effects on the RHS are small imperative programs that can also modify the knowledge base.
Note that external C++ code may be invoked as well, by providing so called user-defined functions to the environment via the C++ API.

The CLIPS inference engine is responsible for the execution of a CLIPS program, initiated by the \emph{(run)} command. It employs a Rete Network~\cite{forgy_rete_1982} to continuously maintain an \emph{agenda} of activated rules. Rules fire sequentially until the agenda is empty.
However, since firing a rule typically modifies the knowledge base and thereby the list of activated rules, the agenda dynamically updates throughout execution.

\begin{figure}
\includegraphics{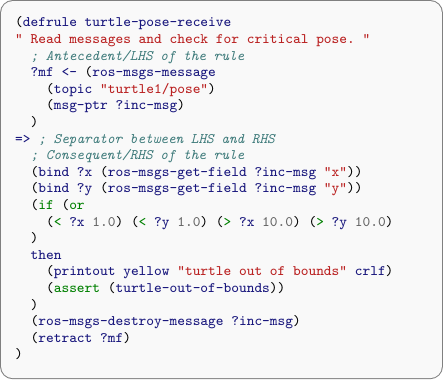}
\caption{Example CLIPS rule, processing a ROS message.}
\label{fig:rule}
\end{figure}

As a running example, consider the ROS turtlesim environment of the ROS beginner tutorial, where a 2D turtle can be moved along a fixed canvas using keyboard input.
Whenever the turtle hits the canvas border, a message indicates that a collision occured.
In order to stop the turtle from colliding with the border, we can monitor the current position and intervene if needed.
A CLIPS rule to detect the turtle leaving a safe area is depicted in \reffig{fig:rule}, where a rule named \emph{turtle-pose-receive} is defined.
It is conditioned on a fact storing a message from the \texttt{turtle1/pose} topic and processes the message, checking for an out-of-bounds violation (asserting a new fact \texttt{(turtle-out-of-bounds)} if needed), and finally deleting the message data and the related fact.
Note that the functions \texttt{ros-msgs-get-field} and \texttt{ros-msgs-destroy-message} are not native to CLIPS, but are provided by the ROS CX (see \refsec{sec:plugins}).

\subsection{\uppercase{PDDL}}
The \acl{PDDL}~\cite{ghallab_pddl_1998} is the de facto standard language for expressing planning problems in artificial intelligence. It separates the specification of a domain, which defines the available objects, predicates, actions and their preconditions and effects, from a problem description, which specifies the initial state and desired goals.
Given such a description, a PDDL planner computes actions (a plan) that transforms the initial state into one that satisfies the goal conditions.

Over time, several versions of PDDL have been introduced, each extending the language with new features. For instance, PDDL 2.1~\cite{fox_pddl21_2003} added support for durative actions and numeric fluents, allowing reasoning about time and continuous resources, while later versions such as PDDL 3.0~\cite{gerevini_deterministic_2009} introduced preferences and state trajectory constraints.

\section{\uppercase{CLIPS-Executive}}
The main objective of this work is to provide a ROS framework for \ac{CLIPS}, allowing users to build knowledge-driven agents in the ROS ecosystem.
The functionality of CLIPS is captured through \emph{environments}.
A CLIPS environment contains all rules and facts that form a CLIPS program and may be accessed trough the \ac{CLIPS} C++ API, e.g., to parse text files containing CLIPS code, run the inference engine on the current fact base or to extend CLIPS by user-defined functions.
It is possible (though rarely needed) to instantiate multiple environments, e.g., in order to maintain multiple possible hypotheses of the current world state.

The core features of the \acl{CX} framework include a bundled installation of CLIPS, ROS node for managing CLIPS environments and a plugin interface that allow users to customize and extend CLIPS environments.
Additionally, predefined plugins provide fundamental capabilities such as executing user code and running the inference engine, as well as seamless interoperability between ROS and CLIPS, enabling the exchange of data via messages, services, and actions directly within CLIPS programs.

\subsection{\uppercase{CLIPS Vendor}}
To support the integration of \ac{CLIPS} into ROS 2, we provide
\ifnopackagenames
a ROS package
\else
the \texttt{clips\_vendor} package
\fi, a CMake-based wrapper around the CLIPS system.
Unlike the original \emph{make}-based build, our vendor package builds dynamic libraries for flexible usage within the ROS 2 ecosystem.
It includes the implementation, command-line interface, and optionally the CLIPSJNI applications, demos, and example programs from the upstream repository.
The library is provided in three forms: a C version (as \ac{CLIPS} is a C application), a C++ version, and a namespaced wrapper that encapsulates the original code to reduce conflicts with other projects.
The \ac{CX} is built on top of this vendor package and makes use of the name\-spaced library to ensure robust integration with ROS 2 and other third-party components.

\subsection{\uppercase{CLIPS Environment Manager}}
\begin{figure}
	\includegraphics[width=\columnwidth]{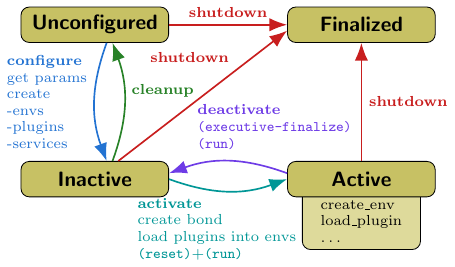}
\caption{\ac{CLIPS} Environment manager ROS node (CX node).}
\label{fig:cx-node}
\end{figure}
The core of the \ac{ROS} \ac{CX} is the environment manager node (also called \ac{CX} node and depicted in \reffig{fig:cx-node}), that manages a \ac{CLIPS} instance, allowing to create and destroy \ac{CLIPS} environments, as well as extending environments by plugins.
Plugins have access to all environments they are loaded into, allowing users to add functionality through the \ac{CLIPS} C++ API.

\begin{figure}
\includegraphics{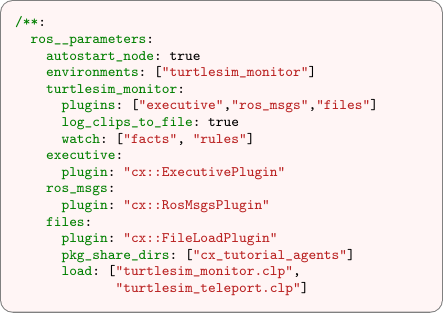}
\caption{Example YAML config for CX node.}
\label{fig:yaml-config}
\end{figure}

The CX node is configured via \ac{ROS} parameters, defining the environments and plugins that should be loaded.
To return to the turtlesim example from \refsec{sec:CLIPS}, where the turtle position needs to be continuously monitored, a suitable configuration is depicted in \reffig{fig:yaml-config}.
The functionality of the plugins are further described in \refsec{sec:plugins}, but essentially, one CLIPS environment takes care of the monitoring, by periodically running the CLIPS inference engine (the \emph{executive} plugin) with monitoring rules defined in files (loaded through the \emph{files} plugin).
In order to send and receive ROS data, the environment is extended by a plugin offering the required functionality (the \emph{ros\_msgs} plugin).
The dynamic parameterization is inspired by the Navigation2 framework~\cite{macenski_marathon_2020}, where it is used in similar fashion to define exchangeable navigation components.

Similarly, the provided node is set up as a ROS lifecycle node with optional bond capabilities, following the approach used in Navigation2 and allowing other nodes to monitor our node's status.
Lifecycle nodes have defined states that give users explicit control over its initialization, configuration, activation, and shutdown.
A bond timer can be used to monitor the node’s liveness and automatically trigger cleanup or recovery if the connection is lost, improving fault tolerance.
Upon configuration, the \ac{CX} node creates all listed environments, instantiates all required plugins and sets up services for managing environments and plugins at runtime.
When the node is activated, the bond becomes active and plugins are loaded in the respective environments. Lastly, each environment is triggered for execution by calling the \ac{CLIPS} functions \texttt{(reset)} and \texttt{(run)}, resulting in executing the CLIPS inference engine on the initial set of facts.

The \ac{CX} node additionally registers a custom log router for all environments.
The router passes log statements to the ROS logging system and optionally also to a file and also supports colored logging statements for shells supporting ANSI escape codes.

\subsection{\uppercase{CLIPS-Executive Plugins}}\label{sec:plugins}
The plugin architecture follows the canonical design pattern in ROS 2, relying on pluginlib\footnote{\url{https://docs.ros.org/en/rolling/Tutorials/Beginner-Client-Libraries/Pluginlib.html}} for discovery and runtime loading.
Plugins inherit from a common base class and follow a well-defined lifecycle: \texttt{initialize()} is called once upon loading, \texttt{clips\_env\_init()} and \texttt{clips\_env\_destroyed()} are invoked once per environment when loading or unloading a plugin, and \texttt{finalize()} is called when the plugin is destroyed.
The system enforces strict ordering of plugin initialization and unloading to ensure safe and predictable integration across multiple CLIPS environments.
Note that the same plugin may also be loaded into mulitple environments.
This allows to reuse plugins if applicable, as well as opening the possibility to bridge between environments via a plugin, if needed.

Through the provided functions, users can access the environment and a mutex for guarding potential asynchronous operations. Direct interactions with the environment during \texttt{clips\_env\_init()} or \texttt{clips\_env\_destroyed()}, as well as code invoked from within an environment (user-defined functions for \ac{CLIPS}) are already guarded through the mutex by the invoking entity.
Every other kind of asynchronous operation accessing \ac{CLIPS} data must be guarded, as \ac{CLIPS} is not thread-safe.
E.g., the \emph{RosMsgsPlugin} described below uses the mutex in order to provide ROS data to the CLIPS environment. Whenever a message is received via a callback, the mutex is acquired in order to assert a corresponding CLIPS fact.

% The API documentation for \ac{CLIPS} is contained in the \ac{CLIPS} \emph{Advanced Programming Guide}. A tutorial for building plugins can be found on the project documentation page \footnote{\url{https://carologistics.github.io/clips_executive/clips_executive/index.html}}.

\paragraph{File Load Plugin}
The main plugin to load any user-code is the \emph{FileLoadPlugin} that simply loads \ac{CLIPS} source files using the \emph{batch*} or \emph{load*} functions of \ac{CLIPS}.
To accommodate for the usual setup of \ac{ROS} projects, it supports the specification of file paths relative to a \ac{ROS} packages share directory.

\paragraph{Executive Plugin}
Continuous Execution of the \ac{CLIPS} engine is enabled through the \emph{ExecutivePlugin}. It attempts to call the \textit{(run)} command of each environment consecutively with a given frequency using a ROS timer (acquiring the corresponding mutex first).
The actual run frequency may be lower, if the agendas cannot be fully processed in time.
An optional empty message to a \ac{ROS} topic can be emitted to measure the actual frequency of engine runs.
Lastly, it can also insert the current \ac{ROS} time as a fact directly and lets CLIPS code retrieve it via appropriate functions.% if the \emph{assert\_time} value is set.

\paragraph{ConfigPlugin}
For convenient parametrization via configuration files, the \emph{ConfigPlugin} enables the parsing and transformation of YAML files into CLIPS facts, allowing adjustments for agents without touching the actual code.

\paragraph{ProtobufPlugin}
The \emph{ProtobufPlugin} allows to send and receive messages encoded via Google Protocol Buffers by wrapping the \emph{protobuf\_comm} library.
The messages are annotated with a message header to annotate message type and size. Basic encryption is supported as well.
This plugin is especially useful for interfacing with other components that are not part of the ROS ecosystem, as Protobuf is widely available on many platforms and has bindings for a wide range of programming languages.
Note that as \emph{protobuf\_comm} is licensed under GPLv2+ license, this plugin follows this license as well.

\paragraph{AmentIndexPlugin}
The remaining collection of plugins eases the integration between \ac{CLIPS} and \ac{ROS}.
The \emph{AmentIndexPlugin} offers bindings to query the ament index via the \emph{ament\_index\_cpp} package. This is primarily useful when handling resources provided by ROS packages. E.g., in order to provide a PDDL domain file to the \emph{PDDL Manager} (see \refsec{sec:PDDL-CLIPS}), a full path is required, which can be retrieved relative to the providing package using the provided CLIPS function \texttt{ament-index-get-package-share-directory}.

\paragraph{RosParamPlugin}
Parameters of the \ac{CX} node can be retrieved directly using \emph{RosParamPlugin}. This is especially useful for passing launch parameters to \ac{CX} agents.

\begin{figure}
\includegraphics{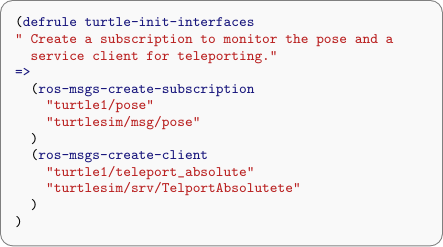}
\caption{Example to create \ac{ROS} interfaces via a CLIPS rule. The rule has an empty antecedent and is therefore fired exactly once.}
\label{fig:ros-msgs-example}
\end{figure}
\paragraph{RosMsgsPlugin}
The \emph{RosMsgsPlugin} manages access to the ROS communication interfaces.
It offers CLIPS functions to publish and subscribe to \ac{ROS} topics, provide and call \ac{ROS} services and to create action clients.
Examples of these functions were also used earlier in the rule for monitoring the turtle position (\reffig{fig:rule}).
\reffig{fig:ros-msgs-example} shows an additional rule for the ongoing example, where the ROS interfaces are used to monitor (via the \texttt{turtle1/pose} topic) and rescue the turtle if needed (via the \texttt{turtle1/teleport\_absolute} service).
The implementation relies on the \ac{ROS} introspection API, which has varying availability in current ROS versions. Its granularity is a result of following the ROS2 C++ API.
E.g., service client introspection was introduced in \textit{Jazzy}, service providers and action clients were added in \textit{Kilted} and action servers are not supported as of yet.

\begin{figure}
	\includegraphics{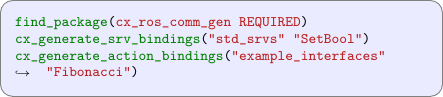}
\caption{Example to generate interface plugins via CMake.}
\label{fig:ros-comm-gen}
\end{figure}

In addition to generic ROS interfaces,
\ifnopackagenames
the \emph{cx\_ros\_comm\_gen} package
\else
a ROS package
\fi
offers convenient CMake macros to generate \ac{CLIPS} bindings per interface, which allow to communicate regardless of the level of introspection support, as the snippet in \reffig{fig:ros-comm-gen} showcases.

\paragraph{Tf2PoseTrackerPlugin}
Lastly, in order to allow reasoning with spatial data, the \emph{Tf2PoseTrackerPlugin} wraps the standard tf2\_ros library to monitor poses from the ROS transform tree by querying information currently stored in the tree periodically via ROS timers.

All the provided plugins have usage examples bundled in 
\ifnopackagenames
a dedicated package,
\else
the \texttt{cx\_bringup} package,
\fi
along with a launch file for starting the \ac{CX} either through composition or directly.

\subsection{\uppercase{Example: Turtlebot Boundary Checks}}
\label{sec:turtle}

To demonstrate the integration of the \ac{CX} with ROS, let us complete the example for monitoring the turtlesim simulator.
The objective is to keep the simulated turtle within a defined safe area by continuously observing its position and applying corrective actions as needed.
This monitoring is realized through a CLIPS environment that interleaves inference with ROS communication.
The setup relies on three \ac{CX} plugins: the \emph{ExecutivePlugin}, \emph{RosMsgsPlugin} and \emph{FileLoadPlugin}.
Together, these plugins establish a self-contained execution environment, in which rules can be repeatedly evaluated against the latest ROS data.
In this example, the turtle’s position is obtained from the ROS topic \texttt{turtle1/pose}.

\begin{figure}
\includegraphics{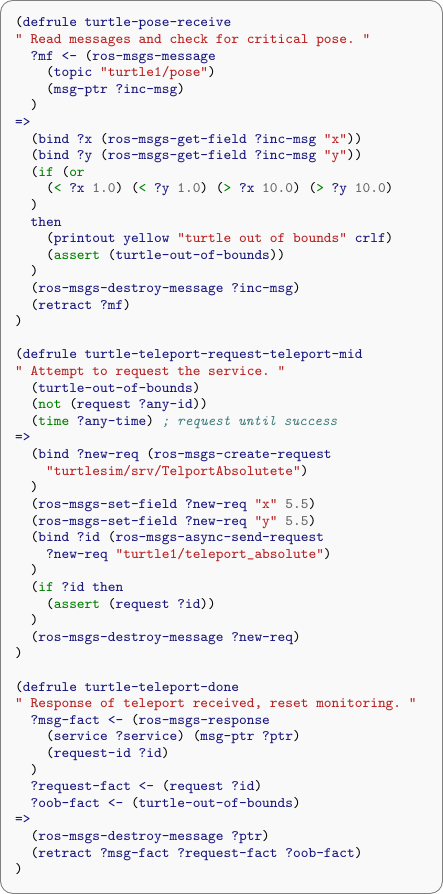}
\caption{Detect out-of-bounds turtle in CLIPS.}
\label{fig:detect-turtle-example}
\end{figure}

The configuration is depicted in \reffig{fig:yaml-config}. The code to open the required ROS interfaces and to observe the turtle pose are shown in \reffig{fig:rule} and \ref{fig:ros-msgs-example}. The only missing piece is a set of rules that requests the teleportation of the turtle to a safe position if needed, for which we give an example in \reffig{fig:detect-turtle-example}.

Whenever the turtle leaves the virtual bounds, the rule asserts the fact \texttt{(turtle-out-of-bounds)}. A complementary set of rules then activates a ROS service client to \texttt{turtle1/teleport\_absolute}, which repositions the turtle to the center of the map.
The client is created and managed by CLIPS rules through the \emph{RosMsgsPlugin}, and requests are retried until the teleportation service acknowledges completion.

This example demonstrates how the \ac{CX} provides a modular architecture for coupling rule-based reasoning with robotic middleware. By configuring a CLIPS environment with the appropriate plugins and rule files, one can achieve continuous monitoring and reactive control.

\section{\uppercase{PDDL Planning with the CLIPS-Executive}}
\label{sec:PDDL-CLIPS}
%this is very important: we whould make it more clear why PDDL. so I added a sentence here
\ac{CLIPS} is excellent for the reactive execution and monitoring that is desirable for building an agent framework. With the provided plugins the \ac{CX} can be used flexibly to build agents that are reactive, thanks to continuous execution of the rule base through the \ac{CLIPS} inference engine. However, this type of approach is less suitable for long-horizon problem solving, which is another critical aspect of modern autonomous robot systems. Achieving complex objectives that require many steps, optimizing resource utilization, and other types of deliberative reasoning are better modeled through \ac{PDDL} planning~\cite{ghallab_pddl_1998}, which allows representing goals, actions, and constraints to generate plans that achieve the desired objectives.
In~\cite{niemueller_clips-based_2018,niemueller_goal_2019}, Niemueller et al. present an integration of \ac{PDDL} into the \emph{Fawkes} \acl{CX} that is both used for planning and for execution.
In their work, a PDDL domain and problem is modeled via \ac{CLIPS} facts, logic for checking conditions and applying effects of actions is encoded via rules and an external planner can be called on demand.%For planning, they transform the CLIPS-based representation to PDDL first and then pass it to a planner.

A core limitation of the \emph{Fawkes} integration of \ac{PDDL} is the need to evaluate action preconditions and effects in \ac{CLIPS}. The provided implementation is rather basic, as advanced PDDL features, such as quantified formulae, conditional effects, numerical fluents and temporal conditions are not supported.
Instead, we propose to utilize existing frameworks for handling PDDL, leveraging their features, modeling capabilities and compliance to the PDDL standard.
In particular, we provide an integration with \ac{UPF}~\cite{micheli_unified_2025}, which can handle a wide set of PDDL standards.

While the \ac{UPF} handles the parsing and solver invocation, our PDDL Manager node wraps this functionality into ROS interfaces that the CLIPS-Executive can control. We translate the results from UPF via ROS into CLIPS facts, enabling the hybrid reactive-deliberative loop without the need to reimplement \ac{PDDL} logic in \ac{CLIPS}.
Additionally, we bundle an existing implementation of the NEXTFLAP planner~\cite{sapena_hybrid_2024} in \ac{UPF}, which can handle classical and temporal PDDL variants.

%\subsection{\uppercase{PDDL Manager}}
These features are provided via the \emph{\ac{PDDL} Manager}, a ROS lifecycle node with bond timer, just like the \ac{CX} node.
It enables the management of multiple parallel instances of PDDL environments, provides services for loading PDDL files (optionally also templated via jinja), managing fluents, objects, functions, goals, checking conditions and applying effects. Each instance may have multiple goals associated to it, a planning action server can be invoked for each goal, allowing to plan concurrently for alternative objects. Planning filters may be used in order to narrow down a planning problem to a subset of actions, objects and fluents. This can be helpful when a detailed execution model is needed which is too verbose for planning.

\begin{figure}
\includegraphics{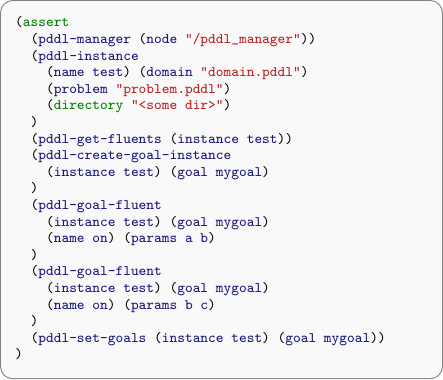}
\caption{Example to create a PDDL instance, getting all current fluents and setting a goal.}
\label{fig:pddl-clips}
\end{figure}

The \emph{PDDL Manager} is then integrated with \ac{CX} agents through ROS services and action servers.
\ifnopackagenames
A dedicated helper package
\else
The \texttt{cx\_pddl\_clips} package
\fi
offers a set of CLIPS rules, which provide additional helpers for tighter integration.
\reffig{fig:pddl-clips} depicts an example snippet in the blocksworld domain, a commonly used toy example. By asserting the depicted facts, a PDDL instance is created by parsing a domain and problem file, the initial list of fluents is retrieved and then a goal is created for stacking a tower via the blocks $a$,$b$ and $c$.

Full examples for using the \emph{PDDL Manager} with or without the help of auxiliary rules of the
\ifnopackagenames
helper package
\else
\texttt{cx\_pddl\_clips} package
\fi
are provided
\ifnopackagenames
\else
in the \texttt{cx\_pddl\_bringup} package,
\fi
along with a launch file for launching the \emph{PDDL Manager} along with the \ac{CX}.
Note that there is also an alternative implementation for integrating \ac{UPF} with ROS\footnote{\url{https://github.com/PlanSys2/UPF4ROS2}}, but it lacks features such as evaluating conditions of actions. Hence, in the following, we describe our UPF integration for ROS through a running example.

\section{\uppercase{Example: Production Logistics}}
\begin{figure*}
	\includegraphics{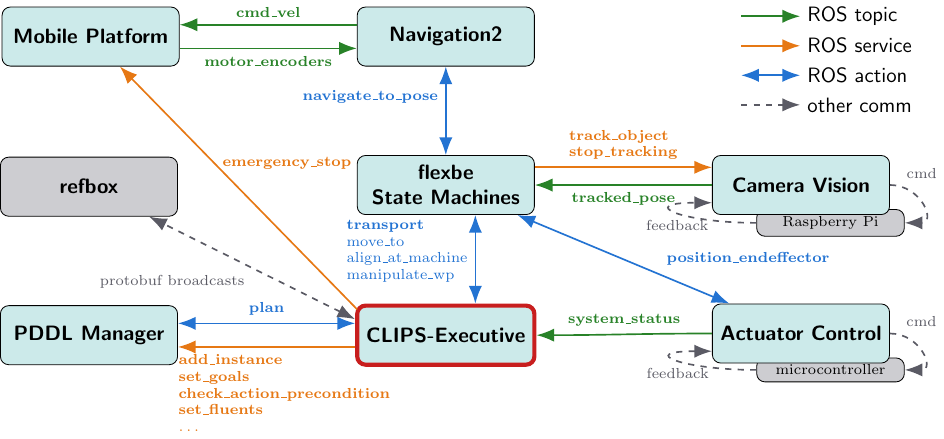}
	\caption{Robotic system overview for the \ac{RCLL} with \ac{CX}-based agent.}
\label{fig:full-system}
\end{figure*}
To illustrate how the \ac{CX} framework can be applied in complex, real-world scenarios, we present a comprehensive example in the context of production logistics.
This case demonstrates the integration of knowledge-based reasoning, task planning, and multi-robot coordination in a dynamic environment, highlighting the practical capabilities of \ac{CX}-based agents.
In the robotics competition \acf{RCLL}~\cite{niemueller_robocup_2015} a team of three
mobile robots have to manufacture dynamically ordered products in a smart factory environment.
They do so by operating a set of static production machines at randomized positions.
The main challenges revolve around finding efficient production plans to maximize the factory output, coordinating the robots for effective cooperation, and reacting to unexpected events and failed robot actions.

% Hofmann et al.~\cite{hofmann_multi-agent_2021} present a distributed CLIPS-based system integrated into Fawkes for the \ac{RCLL}.
% It models objectives as goals, where goals are either transportation tasks (consisting of the actions to pickup a product from a station, driving to a station and placing it down at a station), or machine instructions to process workpieces (consisting of an action to send the instruction and another one to apply the resulting effect). The goals are expanded into plans, with actions mapping to lower-level skills, using a predefined plan library, goals are statically asserted in a tree structure. Each robot spans it's own goal trees, synchronization and locking mechanisms ensure that robots do not perform conflicting or duplicated goals.

Here, we present a novel approach to solve the \ac{RCLL} problem using the ROS 2 \ac{CX}, leveraging temporal planning, a hierarchical PDDL model, and dynamic replanning.
An overview of the resulting system is depicted in \reffig{fig:full-system}. It is built around a centralized \ac{CX}-based agent, which interacts with other ROS components, such as our \emph{PDDL Manager}, the FlexBE behavior engine~\cite{zutell_ros_2022}, and the Navigation2 framework~\cite{macenski_marathon_2020}.
While the agent and full integration with FlexBE are still under development, we focus here on the completed components supporting high-level reasoning and action execution, implemented with the \ac{CX} and the \emph{\ac{PDDL} Manager}.
These modules were deployed at RoboCup 2025 in Salvador, validating their feasibility in a real robotic environment. 
However, this deployment was purely a case-study and test of functionality, not a test of performance.

\subsection{\uppercase{PDDL Action Model}}
\begin{figure}
\includegraphics{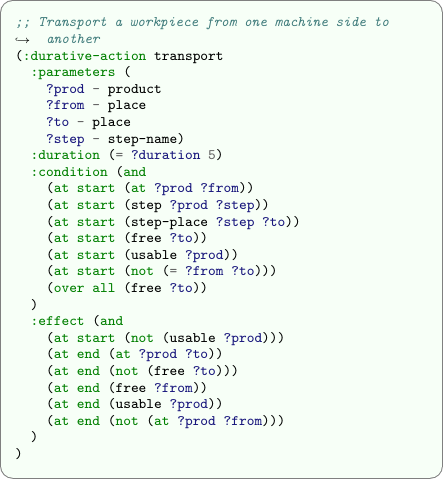}
\caption{PDDL transport action.}
\label{fig:pddl-transport}
\end{figure}

In the \ac{RCLL}, the atomic actions corresponding to low-level skills consist of robot navigation, grasping operations for picking up and placing workpieces, and issuing commands to the static production machines to initiate their respective assembly tasks.
Compared to~\cite{hofmann_multi-agent_2021}, where each atomic action was modeled directly, here the chosen action model is more coarse.
E.g., we use a single \texttt{transport} action to model the change in world state due to the pickup, moving to a destination, and placing down of a workpiece at its target (depicted in \reffig{fig:pddl-transport}).
De Bortoli et al.~\cite{de_bortoli_evaluating_2023} show that a coarse action model is beneficial for high-level planning in the logistics league, enabling the translation of incoming orders into PDDL goals.

Such an abstracted planning domain also allows quick planning for different sets of product orders before committing to plans deemed suitable by some criteria.
However, execution of these coarse actions is still realized through the atomic actions, e.g., navigating or grasping, which may require more detailed parameters, such as target poses or grasping strategies.
These details might depend on the current world state, which may change between planning and execution.
To bridge this gap, we established a hierarchical planning model where the coarse PDDL actions are refined into more detailed plans just before execution for dynamically adapting to the current situation.

%\begin{figure}
%\includegraphics{pddl_mount_cap.pdf}
%\caption{PDDL machine instruction action.}
%\label{fig:pddl-mount-cap}
%\end{figure}

\paragraph{Sub-Actions for Partial Feedback}
As coarse action models aggregate multiple lower-level skills into a single action on the planning level, the conditions and effects of these skills are merged into a single representation as well.
This aggregation restricts the possibility of concurrent execution with other actions, since action effects are only applied at coarse temporal boundaries.
De Bortoli et al.~\cite{de_bortoli_improving_2024} mitigate this limitation by decomposing the coarse \emph{macro actions} into sequences of atomic actions.
However, compound actions such as \texttt{transport} can also be beneficial for the executing layer, as they enable seamless transitions between routines.
For instance, grasping procedures may be started as soon as the object of interest enters range (requiring an early abort of the navigation skill), rather than waiting for the navigation to terminate driving to a predetermined position.

In this work, we retain the abstraction of coarse action models from planning during execution, but enhance them with mechanisms for partial feedback during execution.
Actions are permitted to emit intermediate feedback events, which are mapped to an extra set of actions in PDDL we call sub-actions. 
The corresponding effects are applied incrementally upon receiving feedback, rather than being delayed until the completion of the entire action.

During a \textit{transport} action, once a workpiece has been picked up, the effect of freeing the machine output can be applied immediately, even though the transport action continues until delivery is finalized, if we use such sub-actions to model the effects.
This approach preserves the modeling advantages of coarse actions while enabling more continuous updates to the world state, facilitated by the \ac{CX} agent, which allows us to collect the feedback from multiple sources and define rules for handling \textit{partial feedback}.

%\paragraph{Domain Validation}
%Additional sanity checks can be performed while building the PDDL domain to validate the correctness of modeled sub-actions, to mitigate inconsistent world states during execution.
%Specifically, unit-tests can be automatically derived from the coarse action by creating a planning instance using its preconditions as an initial state, and its effects as a goal.
%If no such plan exists, the sub-actions cannot be considered a faithful decomposition of the original coarse action.

\paragraph{Temporal Planning}
While temporal planners typically produce schedules with precise time points for each action, faithfully adhering to these timestamps during real-world execution is typically infeasible.
Small deviations in action execution or unexpected delays would immediately invalidate the original schedule.
To mitigate this, instead of committing to the exact start and end times for actions in the temporal plan, we relax the plan to a partial ordered one, grouping actions with identical start times for parallel execution. 
During execution, actions within the same group may be executed in any order or even concurrently, as long as their preconditions are satisfied. The next group of actions is then started only after all actions in the current group have completed.

One of the major benefits of sub-actions and early effect application emerges.
Due to our relaxation into groups, the agent can check continuously which preconditions are satisfied and can assign the next possible action to one of the available workers.
We combine this with a lookahead mechanism that considers future actions for early execution, which increases the parallelism of the execution even further.

A potential further improvement would be to consider temporal constraints between actions and analysing their causal links to synthesize a more accurate representation, e.g., as done in~\cite{de_bortoli_improving_2024}, where temporal plans are converted to \ac{STN} for execution.

%Similarly, failed execution attempts can also modelled like this, by having PDDL actions that apply the consequences of a failure.

\subsection{Plan Execution}
Once a plan is generated, it needs to be executed. To this end, the \ac{CX} agent retrieves the plan from the \emph{PDDL Manager} and asserts it as a set of CLIPS facts.
Actions are dispatched according to their order and the preconditions, and assigned to suitable workers for execution.

\paragraph{Worker Assignment} 
Whenever a plan is to be executed, each PDDL action is assigned to a specific worker type.
In our \ac{RCLL} example, we distinguish between the following types of workers:
%\begin{itemize}
robots, which share an identical skill set and are capable of transporting workpieces;
machines, which are controlled by sending instruction messages through the CLIPS agent;
and the agent itself.
%\end{itemize}
When an action is assigned to the agent worker, it is marked for hierarchical planning and refined lazily just before execution.
Its intended effects become goals for the planner, producing a more detailed plan to achieve the task.
For example, in the \ac{RCLL}, there is a machine that can provide material from two sides.
The appropriate side is selected only when the action is executed, depending on whether another robot is already occupying or approaching one of the sides.

Our example highlights our framework's capability of handling multiple mobile robots. In our logistics domain, plans are entirely robot agnostic.
Task assignment to workers is deferred to the execution phase, where any available robot, or other suitable worker, can be assigned to the handle the next action.
% This contrasts with the approach in~\cite{hofmann_multi-agent_2021}, where goals and plans are generated individually for each robot, requiring explicit conflict avoidance mechanisms such as resource locking before a goal can be pursued.
% Furthermore, world model synchronization is unnecessary in our approach, as the centralized architecture maintains a single CLIPS environment that contains all information relevant to all robots.

\paragraph{Action Dispatch} 
Action dispatch is coordinated through the constraints of the partially ordered plan.
When an action becomes applicable for execution (i.e., it is in the currently active group in the relaxed plan), its conditions are checked against the current state through the \emph{PDDL Manager}.
Actions are only executed, if the conditions are currently satisfied, ensuring consistency between the planned actions and the current knowledge base, similar to the implementation of Hofmann et. al.~\cite{hofmann_multi-agent_2021}.

Instead of just considering the current group in the relaxed plan, we also allow actions from a lookahead set to be executed early, if their conditions are already satisfied. These actions are part of the next $n$ groups in the relaxed plan, where $n$ is a configurable parameter.

Actions, where the worker type is a machine or the agent, are handled immediately.
In contrast, actions requiring robots are placed into a pool, from which any available robot may be assigned to he next suitable action for execution.
In order to dispatch an action, another system component is called. E.g., robot actions are handled by a behaviour framework like FlexBe or Navigation2, machine communication is handled from within CLIPS, hierarchical planning is done through the \emph{PDDL Manager}. These components are integrated with the \ac{CX} agent through ROS interfaces, such as topics, services and actions and their feedback is monitored through CLIPS rules.

%\subsection{\uppercase{Error Handling}}
\paragraph{Error Handling} 
%Our CLIPS-based framework is well suited for reactive error handling.
The CLIPS inference engine facilitates reactive handling of unexpected events, as individual rules can be written to handle new information and react to error feedback.

We distinguish two cases: Failures caused by execution itself (e.g., a state machine ending in an error state) or problems observed through general perception independent from robotic action.
The former can be handled by modeling erroneous execution outcomes as PDDL actions with the corresponding effects. Hence, the action feedback maps to the action effect that is applied.
Afterwards, deliberation is necessary to decide whether the current plan remains valid, can be repaired by injecting additional actions (e.g., retrying a failed action) or replanning is needed.

In case action-independent problems arise, the agent first identifies affected goals or constraints.
The CLIPS engine can then trigger rules that update the world model, modify facts, or mark certain objectives as temporarily unreachable.
Depending on the severity and scope of the issue, the system may either adapt the current plan by skipping or reordering actions, or invoke the PDDL planner to generate a new plan that accommodates the updated state of the environment.
This separation between reactive handling of execution errors and deliberate replanning for external events allows the agent to maintain robust and flexible operation in dynamic settings.

\section{\uppercase{Conclusion}}
In this paper, we presented the ROS~2 \acl{CX}, an open-source framework that integrates the CLIPS rule-based reasoning engine into the ROS~2 ecosystem.
Through its plugin architecture and lifecycle-managed environments, the \ac{CX} enables knowledge-driven robotic applications that combine reactive reasoning with deliberative control.
Existing plugins provide seamless integration with ROS and support continuous reasoning via periodic inference cycles, while custom plugins can be used to customize the framework further as needed.

We further presented a CLIPS integration for \ac{PDDL}-based planning via the \acl{UPF}, enabling long-horizon reasoning, temporal planning, and hierarchical action refinement.
Using the \acl{RCLL} as an example, we showed that centralized reasoning supports multi-robot coordination without explicit conflict resolution, while coherent world knowledge and replanning handle execution errors and unforeseen events.

Overall, the \ac{CX} illustrates how \ac{CLIPS} can be effectively combined with ROS~2 to support decision-making that spans reactive to deliberative planning.
%We provide a full example of our turtlebot agent
%\ifnopackagenames
%\else
%in the \texttt{cx\_tutorial\_agents}
%\ifblinded
%package\anonlink.
%\else
%package\footnote{\url{https://index.ros.org/p/cx_tutorial_agents/}}.
%\fi
%\fi

For future work, we plan to expand the feature set of the \ac{CX} to cover additional aspects of high-level reasoning in robotics.
One promising direction is to integrate data-driven approaches, such as reinforcement learning frameworks (e.g., Gymnasium\footnote{https://gymnasium.farama.org/index.html}), with knowledge-based reasoning in the \ac{CX}.
We also aim to improve the user experience of working with CLIPS by developing advanced debugging tools to inspect CLIPS program traces, a language server for \ac{CLIPS}, and a web-based frontend for the \ac{CX}.

\ifblinded
\addtolength{\textheight}{-2cm}
\else
\addtolength{\textheight}{-0cm}
\fi
%\addtolength{\textheight}{-0cm}   % This command serves to balance the column lengths

%%%%%%%%%%%%%%%%%%%%%%%%%%%%%%%%%%%%%%%%%%%%%%%%%%%%%%%%%%%%%%%%%%%%%%%%%%%%%%%%
\section*{\uppercase{Acknowledgements}}
\ifblinded
{\footnotesize
Omitted for anonymization.}
\else
{\footnotesize Tarik Viehmann has been supported by the German Federal Ministry of Research, Technology and Space (BMFTR) under the Robotics Institute Germany (RIG), the Deutsche Forschungsgemeinschaft (DFG, German Research
Foundation) under Germany's Excellence Strategy – EXC-2023 Internet of
Production – 390621612 and Research Training Group 2236 (UnRAVeL).
Daniel Swoboda has been supported by the Alexander von Humboldt Foundation with funds from the Federal Ministry for Education and Research, by the European Research Council (ERC), Grant
agreement No. 885107, and by the Excellence Strategy
of the Federal Government and the State Governments, Germany.
}
\fi
{
\small
\bibliographystyle{apalike}
\bibliography{CX}
}
\end{document}